\documentclass[letterpaper, 10 pt, journal, twoside]{ieeetran}

\IEEEoverridecommandlockouts                              

\usepackage{color,xcolor}
\usepackage{graphicx} 
\usepackage{booktabs}
\usepackage{multirow} 
\usepackage{threeparttable}
\usepackage{colortbl}
\usepackage{makecell}
\usepackage{subfigure}
\usepackage{amsmath} 
\usepackage{amssymb}  
\usepackage[normalem]{ulem}

\definecolor{b}{HTML}{00B0F0}
\definecolor{r}{HTML}{DF0000}
\definecolor{c2}{HTML}{FBD9BD}

\makeatletter
\renewcommand{\maketag@@@}[1]{\hbox{\m@th\normalsize\normalfont#1}}%
\makeatother

\title{
MambaXCTrack: Mamba-based Tracker with SSM Cross-correlation and Motion Prompt for Ultrasound Needle Tracking
}

\author{
Yuelin Zhang, Long Lei, Wanquan Yan, Tianyi Zhang, Raymond Shing-Yan Tang, and Shing Shin Cheng$^{*}$
\thanks{Manuscript received: November, 12, 2024; Revised February, 13, 2025; Accepted March, 21, 2025.}
\thanks{This paper was recommended for publication by Editor Jessica Burgner-Kahrs upon evaluation of the Associate Editor and Reviewers' comments.
Research reported in this work was supported in part by Research Grants Council (RGC) of Hong Kong (T45-401/22-N, CUHK 14217822, and CUHK 14207823) and in part by Innovation and Technology Commission of Hong Kong (ITS/234/21, ITS/233/21, ITS/235/22, and Multi-scale Medical Robotics Center). The content is solely the responsibility of the authors and does not necessarily represent the official views of the sponsors.}
\thanks{Yuelin Zhang is with the Department of Mechanical and Automation Engineering and T Stone Robotics Institute, The Chinese University of Hong Kong, Hong Kong. {\tt\footnotesize ylzhang@mae.cuhk.edu.hk}}
\thanks{Long Lei is with the Department of Computer Science and Engineering, The Chinese University of Hong Kong, Hong Kong. {\tt\footnotesize longlei@cuhk.edu.hk}}
\thanks{Wanquan Yan is currently with the State Grid Xin Jiang Electric Power Co., Ltd. Supervoltage Branch, China. He was previously with the Department of Mechanical and Automation Engineering and T Stone Robotics Institute, The Chinese University of Hong Kong, Hong Kong. {\tt\footnotesize wqyan@link.cuhk.edu.hk}}
\thanks{Tianyi Zhang is with the Medical Physics Graduate Program, Duke Kunshan University, Kunshan, Jiangsu, China. {\tt\footnotesize tz137@duke.edu}}
\thanks{Raymond Shing-Yan Tang is with the Department of Medicine and Therapeutics and Institute of Digestive Disease, The Chinese University of Hong Kong, Hong Kong. {\tt\footnotesize raymondtang@cuhk.edu.hk}}
\thanks{Shing Shin Cheng is with the Department of Mechanical and Automation Engineering, T Stone Robotics Institute, Shun Hing Institute of Advanced Engineering, Multi-Scale Medical Robotics Center, and Institute of Medical Intelligence and XR, The Chinese University of Hong Kong, Hong Kong.
        {\tt\small $^{*}$sscheng@cuhk.edu.hk}}%
\thanks{Digital Object Identifier (DOI): see top of this page.}
}

\begin{document}

\maketitle

\begin{abstract}
Ultrasound (US)-guided needle insertion is widely employed in percutaneous interventions.
However, providing feedback on the needle tip position via US imaging presents challenges due to noise, artifacts, and the thin imaging plane of US, which degrades needle features and leads to intermittent tip visibility.
In this paper, a Mamba-based US needle tracker MambaXCTrack utilizing structured state space models cross-correlation (SSMX-Corr) and implicit motion prompt is proposed, which is the first application of Mamba in US needle tracking.
The SSMX-Corr enhances cross-correlation by long-range modeling and global searching of distant semantic features between template and search maps, benefiting the tracking under noise and artifacts by implicitly learning potential distant semantic cues.
By combining with cross-map interleaved scan (CIS), local pixel-wise interaction with positional inductive bias can also be introduced to SSMX-Corr.
The implicit low-level motion descriptor is proposed as a non-visual prompt to enhance tracking robustness, addressing the intermittent tip visibility problem.
Extensive experiments on a dataset with motorized needle insertion in both phantom and tissue samples demonstrate that the proposed tracker outperforms other state-of-the-art trackers while ablation studies further highlight the effectiveness of each proposed tracking module.

~



\end{abstract}

\section{INTRODUCTION}
\IEEEPARstart{I}{n} various percutaneous intervention procedures, ultrasound (US)-guided needle insertion is commonly adopted in minimally invasive interventions, such as tissue biopsy, tumor ablation, regional anesthesia \cite{liu2019deep}, etc.
As a non-invasive, portable, safe, and cost-effective imaging modality~\cite{richardson2017imaging}, US provides real-time intraoperative imaging of the needle and tissue, thereby minimizing the risk of accidental injury to vessels or critical organs. 
Despite these advantages, {US imaging has an inherent limitation of susceptibility} to \textbf{noise and artifacts} \cite{liu2019deep}, which can obscure or distort the needle tip position.
Furthermore, under the narrow US imaging plane, small structures, such as the needle tip, can \textbf{disappear intermittently}
when they are obscured by anatomical structures or not co-planar with the US imaging plane~\cite{richardson2017imaging}. 
These challenges underscore the necessity for robust and accurate needle tracking to ensure successful insertion procedures under challenging environments.

Prior to the widespread adoption of learning-based needle trackers, traditional methods, such as the statistical filter \cite{mathiassen2016robust} and Gabor filter \cite{kaya2015real}, achieved somewhat satisfactory performance but were hindered by complex workflows and sensitivity to hyper-parameters, failing to address challenging environmental factors in US imaging.
A method based on {a} discriminative correlation filter (DCF) has been proposed in \cite{shen2019discriminative}, but correlation filter-based methods can be susceptible to background distraction and image distortion \cite{yan2023learning}.
{As a result, this DCF method has also been integrated with an optical tracking system for higher accuracy \cite{che2024improving}, but the deployment of an optical tracking system is a cumbersome constraint.}
Recently, deep learning methods based on convolutional neural networks (CNN) and transformers \cite{vaswani2017attention,zhang2024unified} have gained popularity in tracking tasks \cite{li2019siamrpn++,cui_mixformerv2_2023,zhang2024motion}, including US needle tracking \cite{yan2023visual,yan2024task}.
Mwikirize et. al. proposed a US needle tracker with a two-step structure based on a fully convolutional network and {a} region-based CNN \cite{mwikirize2018convolution}.
A paradigm utilizing digital subtraction is proposed in \cite{mwikirize2019learning}, which augments tip features to enhance visibility prior to tracking.
However, these two methods have non-end-to-end structures that require multiple steps to localize the needle tip, potentially affecting robustness and accuracy.
In addition, since the needle tip is often obscured or distorted under artifacts and noise, some other methods perform needle shaft segmentation before localizing the needle tip \cite{wijata2024needle,hui2023ultrasound}. 
{While a segmentation mask of the needle shaft can offer useful cues for determining the axial position of the needle tip, this two-stage workflow may inadvertently introduce accumulative errors and discrepancies. }
Extra data of segmentation mask is also required to train segmentation models, leading to additional obstacles for model deployment.

As an effective structure, cross-correlation (X-Corr) has been widely adopted in end-to-end learning-based trackers. 
It measures the similarity between a reference template and a search region by performing convolution with sliding windows. The target location is then obtained from the induced similarity score.
Following this diagram, many trackers based on X-Corr have been proposed \cite{li2019siamrpn++,chen2022siamban,guo2020siamcar,yu2020deformable}.
However, the existing convolutional X-Corr has a limited modeling range constrained by the kernel size. 
It cannot learn distant semantic features (e.g. needle shaft) which are important for US needle tracking, since local information can unpredictably become unreliable due to degradation by noise and artifacts.

In addition to the challenging environment with noise and artifacts, the intermittently visible needle tip poses another problem that hinders accurate needle tracking. This issue can be caused by deviation of {the} US imaging plane or obstruction by anatomical structures \cite{kimbowa2024advancements}.
Mwikirize et. al. proposed a single-shot needle tracker by integrating historical frames to enhance needle tip features, addressing scenarios where the tip is imperceptible or the shaft is invisible \cite{mwikirize2019single}.
Although it integrates historical information, its feature pre-enhancement can unexpectedly shift the original latent features, causing potential inherent information loss. 
Integrating historical needle motion into visual tracking presents another approach, since motion information can serve as an effective non-visual prompt to prevent tracking failure when the needle tip is invisible.
A motion prediction module is integrated with a visual tracker in \cite{yan2023learning}, yet it is constrained by its explicit motion prediction that poses challenges on generalizability when encountering motion from unseen domains.

Mamba \cite{gu2023mamba} has recently drawn considerable attention and is being applied in real-world tracking tasks \cite{wang2024trackingmamba,lai2024mambavt,zhang2024survey}.
Based on the structured state space models (SSMs) \cite{kalman1960new}, Mamba has a computationally efficient long-range lossless modeling capability with its selective scan mechanism. 
{Leveraging this, Mamba-based trackers can efficiently model long-range information}, such as aggregating a video-level template set \cite{lai2024mambavt} or integrating historical frames \cite{wang2024trackingmamba}.
It should be noted that a transformer-based tracker can hardly achieve long-range modeling under similar model size and complexity since {the self-attention mechanism in transformers has quadratic time complexity and memory requirement with respect to the sequence length \cite{vaswani2017attention}}.
 It also requires positional encoding that may not effectively capture lossless long-range dependencies compared to Mamba, which is designed specifically for such tasks.
Thus, Mamba usually outperforms transformer-based methods under similar model size and complexity \cite{gu2023mamba}.
Since no Mamba-based tracker has yet been proposed for US needle tracking, {developing one remains an open research area}.

To address the aforementioned challenges of noise, artifacts, and intermittent visibility in US needle tracking, {in this work, \textbf{MambaXCTrack}, a Mamba-based US needle tracker utilizing SSM cross-correlation (\textbf{SSMX-Corr}) and an \textbf{implicit motion prompt}, is proposed.}
To the best of our knowledge, it is the first time a Mamba-based tracker has been adopted in US needle tracking.
It is also the first time that cross-correlation is implemented with SSM.
Leveraging the long-range modeling capability of SSMs, SSMX-Corr
enables global search and long-range modeling of distant semantic features.
When the needle tip feature is degraded by noise and artifacts, SSMX-Corr avoids tracking failure by implicitly learning distant semantic {features} that potentially {come} from visual cues like {the} needle shaft, rather than by explicitly performing segmentation on the needle shaft like existing methods \cite{wijata2024needle,hui2023ultrasound}.
By further integrating the proposed cross-map interleaved scan (CIS),
SSMX-Corr enjoys global search without losing local pixel-wise interaction between search and template maps to keep positional inductive bias, while existing convolutional X-Corr models \cite{li2019siamrpn++,bertinetto2016fully} only consider local modeling.
To address the intermittent visibility of the needle tip, {an} implicit low-level motion descriptor is introduced as a non-visual prompt in addition to visual features that can unpredictably become unreliable.
Different from \cite{yan2023learning} that trains an external motion predictor to explicitly predict the future motion, the proposed workflow preprocesses motion to obtain the low-level motion descriptor,
which is then implicitly
integrated with visual features.
This implicit low-level motion integration introduces image-agnostic raw motion and ensures an end-to-end network to enhance training stability and tracking robustness.
Extensive evaluations on a dataset of motorized needle insertions in both phantom and animal tissue demonstrate MambaXCTrack's superior performance compared to state-of-the-art methods.
The main contributions are fourfold:
\begin{itemize}
\item SSMX-Corr improves existing cross-correlation with SSM by globally searching and {modeling long-range} distant semantic features, thus learning potential distant visual cues. This represents the first effort to adopt Mamba effectively for US needle tracking.
\item CIS is proposed to provide SSMX-Corr with local pixel-wise interaction and positional inductive bias in addition to global search to enhance overall tracking performance.
\item {An} implicit low-level motion descriptor is adopted to provide a non-visual prompt, thus leveraging motion information to address the challenge of intermittent needle visibility to achieve robust and consistent tracking.
\item The proposed tracker achieves {state-of-the-art (SOTA)} performance on both phantom and tissue experiments. Further ablation studies show the effectiveness of the proposed modules.
\end{itemize}
\section{Methodology}

\begin{figure*}
\centering
\includegraphics[width=\textwidth]{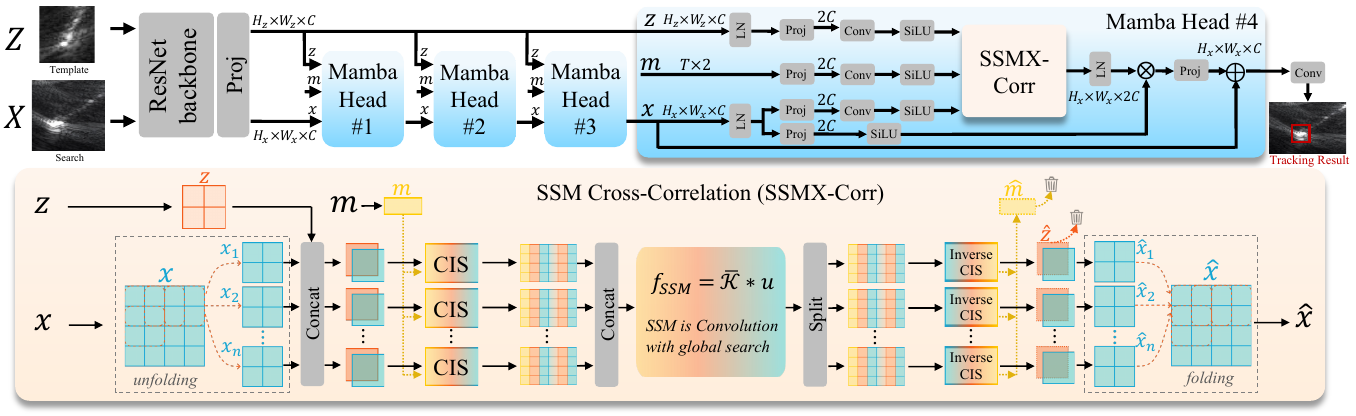}
\vspace{-5mm}
\caption{
Structure overview of the proposed MambaXCTrack. The ResNet backbone is cascaded with four Mamba heads. Each Mamba head has the same structure. $z$ and $x$ are the embedding features of template $Z$ and search $X$.
}
\label{fig_overview}
\vspace{-2mm}
\end{figure*}

\subsection{MambaXCTrack}
The overview of MambaXCTrack is shown in Fig.~\ref{fig_overview}.
{A ResNet-50 network~\cite{he2016deep} with 50 layers is adopted as the backbone.}
The template map $Z$ and the search map $X$ first go through the backbone, from which the embedded features $z\in \mathbb{R}^{H_z \times W_z \times C}$ and $x\in \mathbb{R}^{H_x \times W_x \times C}$ ($C=128$) are obtained through a cascaded linear projection.
In the following Mamba head, after going through cascaded layer normalization, linear projection (in the dimension of $2C$), convolution (kernel size $3\times3$, stride 1), and SiLU activation \cite{elfwing2018sigmoid}, $x$, $z$, and the low-level motion descriptor $m$ are transmitted to SSMX-Corr.
For each Mamba head, $x$ is serially passed to the next head for a better feature representation, yet $z$ and $m$ are concurrently transmitted by each head to avoid error accumulation on the original template and raw motion.

In SSMX-Corr, $x$ is first \textit{unfolded} with a stride $(\frac{H_z}{2}, \frac{W_z}{2})$ to obtain submaps $x_i$ ($i\in \{1,2,...,n\}$) in the same size $H_z\times W_z$ with $z$.
The submaps $x_i$, $z$, and $m$ are scanned respectively by the proposed CIS, then modeled by SSM after being concatenated.
After SSM modeling, the aggregated feature map is scanned inversely to separate $\hat{z}$, $\hat{x_i}$, and $\hat{m}$. $\hat{z}$ and $\hat{m}$ are removed.
The stage output $\hat{x}$ is then obtained by \textit{folding} all $\hat{x_i}$.
The tracking prediction is then obtained from the final $\hat{x}$ by a convolution prediction head.

\subsection{Cross-map Interleaved Scan (CIS)}
\begin{figure}
\centering
\includegraphics[width=0.9\linewidth]{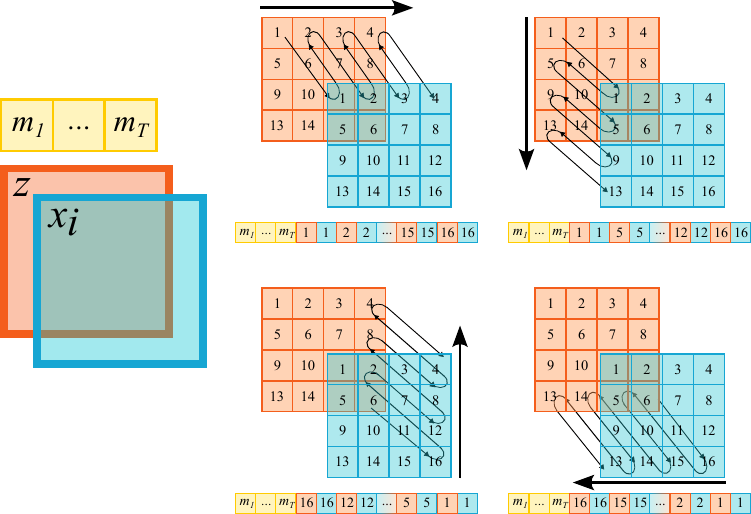}
\caption{
A demonstration of the cross-map interleaved scan (CIS). CIS is performed in four directions, from which four sequences are regrouped separately. 
Assuming $z$ and $x_i$ are all in $4\times4$ with elements numbered $1\sim16$, the induced four sequences with $m$ concatenated are shown.
}
\label{fig_cis}
\vspace{-1mm}
\end{figure}

There exist many scanning paradigms for Mamba \cite{zhang2024survey}. However, there has yet to be a scanning method designed for tracking tasks to enhance the interaction between the template $z$ and search $x$.
The existing cross-correlation-based methods estimate the cross similarity with convolution, which possesses pixel-level interaction and positional inductive bias in nature.
To perform SSM-based cross-correlation, although the model enjoys lossless long-range modeling of semantic information, the local pixel-level relationships between $z$ and $x$ can be unexpectedly neglected if they are simply concatenated together.
In this work, before SSMX-Corr, the CIS is adopted to enhance pixel-wise interaction between $z$ and $x$ to better adapt to SSM-based cross-correlation.
As shown in Fig.~\ref{fig_cis}, CIS receives a template $z$, a submap $x_i$, and a motion descriptor $m$.
Since the SSM modeling is unidirectional, four-directional scanning is adopted.
For scanning in each direction, $z$ and $x_i$ are scanned pixel-by-pixel alternatively as demonstrated in Fig.~\ref{fig_cis}.
Four sequences are then induced, each with the motion descriptor $m$ concatenated at the beginning.
By performing CIS, the local pixels from template and search maps are regrouped to be adjacent, allowing the SSMX-Corr to be performed without losing local interaction and positional inductive bias.

\subsection{SSMX-Corr: SSM is Convolution with Global Search}
The regrouped sequences from CIS are then processed by SSM Cross-Correlation (SSMX-Corr).
The X-Corr in existing trackers \cite{li2019siamrpn++,bertinetto2016fully} calculates the similarity map between template and search maps with convolution.
This convolutional X-Corr operation $f_{conv}$ is given by
\begin{equation}
    f_{conv}(z, x) = z * x,
\end{equation}
where $*$ denotes the convolution.
Since convolutional X-Corr is a local operation, it only models local pixel-wise interaction yet fails to consider interrelationship between distant features.
When local features {from the small needle tip area are} degraded by severe noise and artifacts, X-Corr can {lose} tracking easily without seeking hints from distant semantic features.
Searching globally for potential semantic visual cues then turns out to be a promising paradigm inspired by segmentation-based needle {trackers} \cite{wijata2024needle,hui2023ultrasound}, which obtain the needle's axial direction by segmentation.

To introduce long-range modeling capability into X-Corr, SSMX-Corr is proposed.
SSMs are based on a continuous linear time-invariant (LTI) system that can be formulated as a linear ordinary differential equation (ODE) as follows
\begin{equation}
\begin{split}
    & h^\prime(t)=\textbf{A}h(t)+\textbf{B}u(t), \\
    & y(t)=\textbf{C}h(t),
    \label{eq_ode}
\end{split}
\end{equation}
where $\textbf{A}\in \mathbb{R}^{N\times N}$ {(state matrix)}, $\textbf{B}\in \mathbb{R}^{N\times 1}$ {(input matrix)}, and $\textbf{C}\in \mathbb{R}^{1\times N}$ {(output matrix)}. 
{These two equations together} map the input sequence $u(t)\in\mathbb{R}$ to output $y(t)\in\mathbb{R}$ through {linear transformation of the} latent states $h(t)\in\mathbb{R}^{N\times 1}$. 
{To integrate the continuous system described in Eq.~\ref{eq_ode} into a digital system for deep learning integration, it needs discretization first.}
$\Delta$ is introduced as the step size at timescale. 
Eq.~\ref{eq_ode} is then discretized as
\begin{align}
\begin{split}
    & h_t=\bar{\textbf{A}}h_{t-1}+\bar{\textbf{B}}u_t, \\
    & y_t=\textbf{C}h_t,
    \label{eq_ode_disc}
\end{split}
\end{align}
where the discritized matrices $\bar{\textbf{A}}$ and $\bar{\textbf{B}}$ are given by $\bar{\textbf{A}}=exp({\Delta}\cdot \textbf{A})$ and $\bar{\textbf{B}}=({\Delta}\cdot \textbf{A})^{-1}(exp({\Delta}\cdot\textbf{A})-I)\cdot {\Delta}\textbf{B}$.
During the discretization, {the selective scan mechanism \cite{gu2023mamba} is adopted to learn input-dependent parameters $\textbf{B},\textbf{C},\Delta$ from input $x\in\mathbb{R}^{L\times C}$, where $L$ is the sequence length, $C$ is the dimension.
Parameters $\textbf{B},\textbf{C},\Delta$ are given by $\textbf{B}=\text{Linear}(x)\in\mathbb{R}^{L\times N}$, $\textbf{C}=\text{Linear}(x)\in\mathbb{R}^{L\times N}$, $\Delta=\text{Softplus}(\widetilde{\Delta}+\text{Linear}(x))\in\mathbb{R}^{L\times C}$, where $\widetilde{\Delta}$ is a trainable parameter.
}

By reformulating Eq.~\ref{eq_ode_disc} {\cite{gu2021combining}}, the closed form of the output $y_k$ at every time step $k$ can be derived as
\begin{align}
\begin{split}
    y_k=&\textbf{C}(\bar{\textbf{A}})^k\bar{\textbf{B}}u_0 + \textbf{C}(\bar{\textbf{A}})^{k-1}\bar{\textbf{B}}u_1 + ... \\
        &+ \textbf{C}\bar{\textbf{A}}\bar{\textbf{B}}u_{k-1} + \bar{\textbf{B}}u_k.
    \label{eq_yk}
\end{split}
\end{align}
Then $y$ can be formulated as the result of a convolution {by extracting the coefficients into the SSM kernel $\bar{\mathcal{K}}$}
\begin{align}
\begin{split}
    y &= \bar{\mathcal{K}}*u, \\
    \bar{\mathcal{K}}&=(\textbf{C}\bar{\textbf{A}}^j\bar{\textbf{B}})_{j\in[L]}\in \mathbb{R}^L \\
                       &= (\textbf{C}\bar{\textbf{B}},\textbf{C}\bar{\textbf{A}}\bar{\textbf{B}},...,\textbf{C}\bar{\textbf{A}}^{L-1}\bar{\textbf{B}}).
    \label{eq_k}
\end{split}
\end{align}

In this paper, input $u$ is the scanned aggregation of $m,z,x$ derived from CIS, which is defined by $u=\underset{i=[1,n]}{Cat}(CIS(z,x_i,m))$.
The proposed SSMX-Corr $f_{SSM}$ can then be written as
\begin{equation}
    f_{SSM}(z,x,m) = \bar{\mathcal{K}}*u = \bar{\mathcal{K}}*\underset{i=[1,n]}{Cat}(CIS(z,x_i,m)),
\end{equation}
which is a convolution between the concatenated map from scanned aggregation $CIS(z,x_i,m)$ and the kernel $\bar{\mathcal{K}}$.
{Note that $\bar{\mathcal{K}}$ comes from the parameterization of input ($z$, $x$, $m$)}.
Thus, SSMX-Corr can be regarded as a \textit{self-convolution} of $z$, $x$, $m$, yet it enjoys SSM's long-range modeling capabilities.
With SSMX-Corr, tip tracking can benefit from \textit{global search} between $x$ and $z$.
Distant semantic features from the needle shaft can be learned to guide tracking.
The sequence modeling diagram in SSMX-Corr also facilitates better motion-vision information fusion than existing convolutional X-Corr.

\subsection{Implicit Low-level Motion Descriptor}
There exist some US needle trackers \cite{yan2023learning} that improve tracking performance by {incorporating} motion information. 
However, {explicit motion integration by including a motion predictor makes it a non-end-to-end model, posing potential harm to its generalizability and training stability.}
{An} implicit low-level motion descriptor is introduced in this paper for robust tracking when the needle tip is intermittently invisible.
Given a set of historical bounding boxes $B=\{\beta_1,\beta_2,...,\beta_t\}$ ($t$ is the time index), where $\beta_t=(w_t, h_t, cx_{t}, cy_{t})$ is the bounding box defined by its width $w_t$, height $h_t$, and coordinate of the top-left corner $(cx_{t}, cy_{t})$, the low-level motion descriptor $m_t$ is constructed using the local displacement, given by $m_t=(cx_{t}-cx_{t-1}, cy_{t}-cy_{t-1})=(\Delta cx_{t}, \Delta cy_{t})$.
{The motion descriptor $m_t$ is in pixels units. 
}
This low-level local displacement smooths out the distraction from absolute information at the picture level, benefiting the model inferencing and generalizability.
To store the historical motion during tracking, a motion queue at a length $T$ is built with a FIFO (first-in-first-out) strategy. 
Note that for queue storage, the motion sequence is not sampled by time to keep the original motion step.
The sequence of $m$ is then formulated as $\{m_1,m_2,...,m_T\}$ in the size of $T\times 2$, which is concatenated with $z$ and $x$ in SSMX-Corr for implicit information fusion.

\section{Experiments and Results}
\subsection{Experimental Setup and Dataset Collection}

\begin{figure}
\centering
\includegraphics[width=1.0\linewidth]{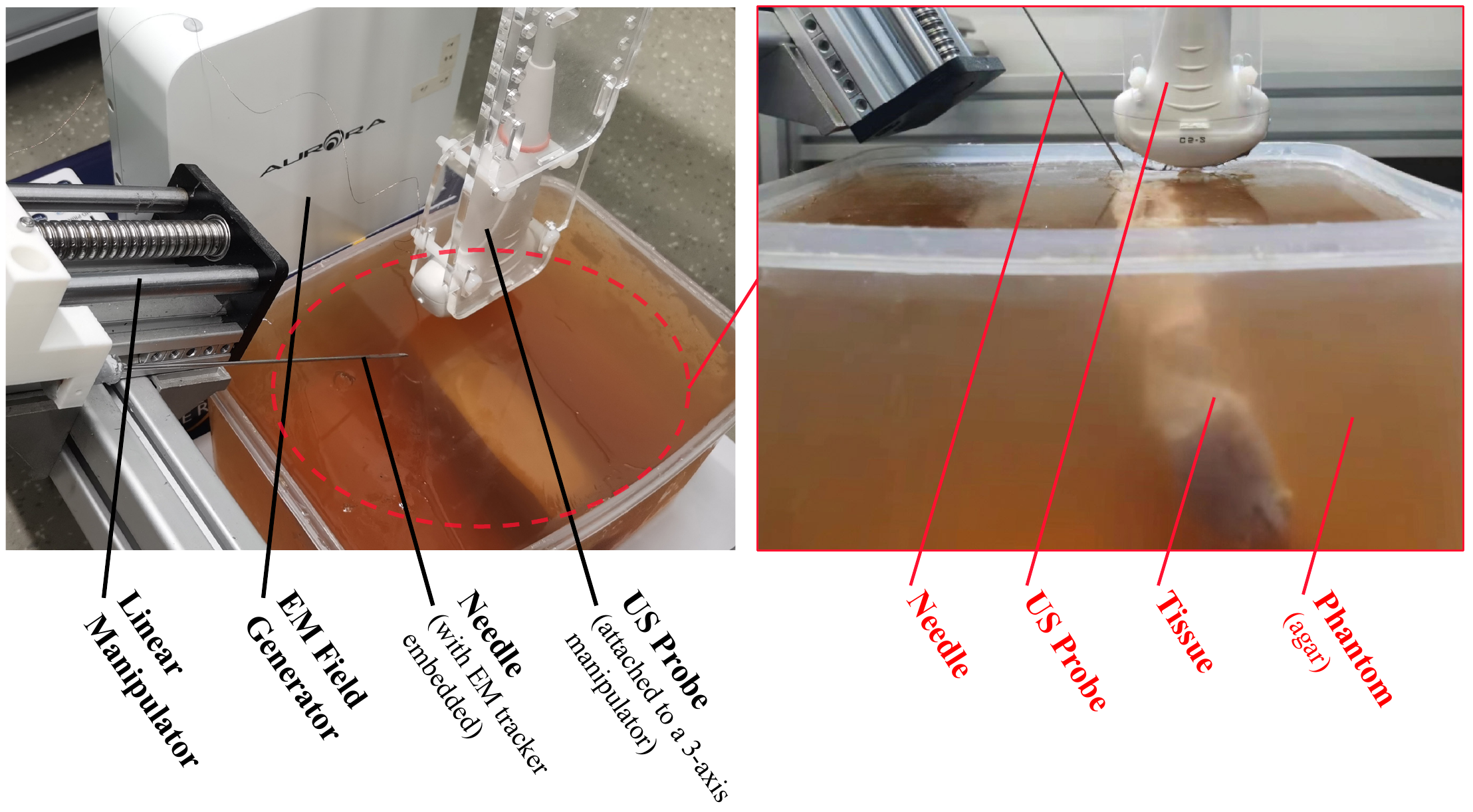}
\caption{Experimental setup. The in-plane case (imaging plane aligned parallel with needle trajectory) is shown as an example.
}
\label{fig_exp}
\vspace{-2mm}
\end{figure}

\begin{table*}
\begin{center}
\caption{
Evaluation results on motorized insertion in phantom and tissue. The methods with the best and the second best performance are noted in {\color{r}red} and {\color{b}cyan} color. All three metrics are reported in percentage (\%).
}
\label{tab_exp}
\setlength{\tabcolsep}{1.1mm}{
\scalebox{0.91}{
\begin{tabular}{lccccccccc|ccccccccc|ccc}
\toprule
\multirow{3}{*}{Method} & \multicolumn{9}{c|}{Phantom}  & \multicolumn{9}{c|}{Tissue}  & \multicolumn{3}{c}{\multirow{2}{*}{Mean}} \\
& \multicolumn{3}{c}{In-plane (static)}  & \multicolumn{3}{c}{In-plane (moving)} & \multicolumn{3}{c|}{Out-of-plane} & \multicolumn{3}{c}{In-plane (static)}  & \multicolumn{3}{c}{In-plane (moving)} & \multicolumn{3}{c|}{Out-of-plane} \\
\cmidrule(lr){2-22}
& \cellcolor{c2!50}AUC & \cellcolor{c2!50}$\textit{P}_{\textit{norm}}$ & \cellcolor{c2!50}$\textit{P}$ & \cellcolor{c2!50}AUC & \cellcolor{c2!50}$\textit{P}_{\textit{norm}}$ & \cellcolor{c2!50}$\textit{P}$ & \cellcolor{c2!50}AUC & \cellcolor{c2!50}$\textit{P}_{\textit{norm}}$ & \cellcolor{c2!50}$\textit{P}$ & \cellcolor{c2!50}AUC & \cellcolor{c2!50}$\textit{P}_{\textit{norm}}$ & \cellcolor{c2!50}$\textit{P}$ & \cellcolor{c2!50}AUC & \cellcolor{c2!50}$\textit{P}_{\textit{norm}}$ & \cellcolor{c2!50}$\textit{P}$ & \cellcolor{c2!50}AUC & \cellcolor{c2!50}$\textit{P}_{\textit{norm}}$ & \cellcolor{c2!50}$\textit{P}$ & \cellcolor{c2!50}AUC & \cellcolor{c2!50}$\textit{P}_{\textit{norm}}$ & \cellcolor{c2!50}$\textit{P}$  \\
\midrule
SiamRPN++ \cite{li2019siamrpn++} & 50.7 & 68.5 & 72.9 & 43.1 & 41.4 & 68.8 & 64.5 & 79.0 & \color{b}84.1 & 29.9 & 30.4 & 50.1 & 44.3 & 32.8 & 81.0 & 66.0 & 85.9 & 92.0 & 48.8 & 55.7 & 73.6 \\
SiamCAR \cite{guo2020siamcar} & 60.0 & 79.6 & 85.5 & 51.7 & 53.0 & 85.4 & 62.1 & 80.5 & 83.9 & 42.8 & 51.2 & 75.1 & 46.4 & 34.1 & 80.2 & 71.3 & 90.3 & 98.1 & 55.5 & 65.0 & 84.7  \\
SiamBAN \cite{chen2022siamban} & 70.5 & 86.6 & 93.5 & 66.0 & 79.3 & \color{r}100.0 & 74.4 & 96.0 & \color{r}100.0 & \color{b}53.2 & \color{b}62.8 & 79.4 & 56.4 & 63.9 & 81.0 & 66.6 & 83.1 & 87.9 & 64.6 & 78.6 & 90.6  \\
SiamAttn \cite{yu2020deformable} & 62.3 & 87.1 & 90.4 & 72.5 & 95.0 & \color{b}99.9 & 74.7 & 97.8 & \color{r}100.0 & 40.4 & 53.5 & 64.6 & 51.6 & 64.1 & \color{b}90.4 & 69.2 & 93.1 & 98.3 & 62.1 & 82.0 & 90.2  \\
Yan et.al. \cite{yan2023learning} & 46.1 & 81.0 & 93.6 & 41.7 & 66.7 & 88.9 & 54.8 & 65.1 & \color{r}100.0 & 24.3 & 36.3 & 58.2 & 29.0 & 39.6 & 64.1 & 45.4 & 83.1 & \color{r}99.9 & 40.1 & 62.9 & 84.3  \\
STMTrack \cite{fu2021stmtrack} & 38.1 & 90.2 & 98.8 & 42.4 & 68.5 & 98.1 & 42.3 & 97.2 & \color{r}100.0 & 42.8 & 61.9 & 79.5 & 37.8 & 50.2 & 72.8 & 51.7 & 92.3 & 97.6 & 42.4 & 76.8 & 91.7  \\
SwinTrack \cite{lin2022swintrack} & \color{b}73.8 & 92.6 & \color{b}99.8 & 65.5 & 76.6 & \color{r}100.0 & 73.2 & 96.0 & \color{r}100.0 & 52.2 & 61.2 & \color{b}79.7 & 56.0 & 64.5 & 81.2 & \color{b}75.4 & 93.0 & 98.5 & 66.1 & 80.6 & \color{b}93.6  \\
MixFormerV2 \cite{cui_mixformerv2_2023} & 71.6 & \color{b}93.4 & 97.6 & \color{b}73.6 & \color{b}96.0 & \color{b}99.9 & \color{r}78.4 & \color{b}97.9 & \color{r}100.0 & 44.2 & 56.3 & 66.5 & \color{r}60.9 & \color{r}75.2 & \color{r}92.6 & 74.5 & \color{b}94.0 & 99.5 & \color{b}67.0 & \color{b}85.6 & 92.5  \\
MambaXCTrack & \color{r}79.2 & \color{r}96.2 & \color{r}99.9 & \color{r}76.0 & \color{r}96.8 & \color{r}100.0 & \color{b}76.3 & \color{r}98.1 & \color{r}100.0 & \color{r}54.8 & \color{r}66.1 & \color{r}83.0 & \color{b}59.4 & \color{b}73.6 & \color{r}92.6 & \color{r}76.4 & \color{r}94.4 & \color{b}99.6 & \color{r}70.8 & \color{r}87.9 & \color{r}95.9  \\
\bottomrule
\end{tabular}
}
}
\end{center}
\vspace{-3mm}
\end{table*}

\begin{table}
\begin{center}
\caption{
Average errors and standard deviations (mm) of the tracking, which are computed over the testing set with 1747 samples. Inference speed is reported in FPS.
}
\label{tab_mm}
\setlength{\tabcolsep}{2.0mm}{
\scalebox{1.0}{
\begin{tabular}{lcccl}
\toprule
Method & Phantom & Tissue & Mean &FPS\\
\midrule
SiamRPN++ \cite{li2019siamrpn++} & 4.11$\pm$3.05 & 3.21$\pm$2.58 & 3.70$\pm$2.83  & 27.9\\
SiamCAR \cite{guo2020siamcar} & 2.26$\pm$1.55 & 2.23$\pm$1.37 & 2.24$\pm$1.47  & 48.3\\
SiamBAN \cite{chen2022siamban} &  0.70$\pm$0.54 & 2.09$\pm$1.55 & 1.33$\pm$1.01  & 30.7\\
SiamAttn \cite{yu2020deformable} &  0.63$\pm$0.67 & 1.62$\pm$1.33 & 1.08$\pm$0.98  & 33.2\\
Yan et.al. \cite{yan2023learning} & 0.98$\pm$0.94 & 2.83$\pm$2.21 & 1.84$\pm$1.53  & 24.0\\
STMTrack \cite{fu2021stmtrack} & 0.41$\pm$0.27 & 2.32$\pm$1.52 & 1.28$\pm$0.86  & 22.1\\
SwinTrack \cite{lin2022swintrack} & 0.37$\pm$0.17 & 1.54$\pm$1.53 & 0.90$\pm$0.81  & \color{b}59.5\\
MixFormerV2 \cite{cui_mixformerv2_2023} & \color{b} 0.26$\pm$0.25 & \color{b} 1.37$\pm$1.51 & \color{b} 0.76$\pm$0.84  & \color{r}81.0\\
MambaXCTrack & \color{r} 0.22$\pm$0.14 & \color{r} 0.49$\pm$1.01 & \color{r} 0.34$\pm$0.55  & 34.9 \\
\bottomrule
\end{tabular}
}
}
\end{center}
\vspace{-5mm}
\end{table}

The experimental setup is shown in Fig.~\ref{fig_exp}.
Ultrasound imaging was performed with a Verasonics Vantage 32 LE US machine with a Mindray C5-2 US probe.
{This probe has a transmit frequency of 3.5 MHz with 96 transducer elements. During imaging, a plane wave sequence with an imaging depth of 150 mm was utilized, and the sampling frequency was set to 25 MHz.}
An electromagnetic (EM) localization system (Aurora, NDI Inc.) was adopted to collect the needle tip's position as the ground truth, using an 18-gauge needle with a 5-DoF EM tracker embedded on its tip.
The needle was attached to a linear manipulator for motorized insertion.
The US probe was installed on a 3-axis manipulator to move the probe for needle visualization when needed.
The phantom was made of agar of 3\% mass concentration and the tissue was a slice of fresh pork.
Silica powder was added to agar to simulate speckles in real anatomical tissue.

During the experiment, the data of the phantom-only experiment and tissue experiment were collected separately regarding three cases, i.e., in-plane (static), in-plane (moving), and out-of-plane \cite{kimbowa2024advancements}.
For cases of in-plane (moving) and out-of-plane, the probe moved at the same speed as the horizontal velocity of needle insertion.
For in-plane (static), only the needle moved.
The needle insertion was performed with three angles ($0^\circ$, $30^\circ$, and $60^\circ$) and three velocities ($0.4$, $1$, and $2$ mm/s) for each case.
For each pair of angle and velocity, at least 6 procedures were conducted.
Each needle insertion length is more than 120 mm to ensure a thorough procedure.
108 procedures have been done in total, including 51 in tissue and 57 in phantom, which results in 108 videos ($492\times856$, 30 FPS) with 87330 frames.
{The pixel resolution of each frame ($492\times856$) corresponds to a field-of-view of 150 mm $\times$ 260 mm.}
The ground truth of the needle tip position was acquired by the EM tracking system, which has been verified to have a satisfactory RMSE of 0.76 mm in our environment.

To train the model, the videos were first sampled by 10 to get 8733 images to remove redundancy.
{The dataset was then split into training (6113 samples), validation (873 samples), and testing (1747 samples) sets in the ratio 7:1:2.}
A video/setup-level data splitting was conducted to ensure images from the same video were not divided into different sets, and videos with different experiment setups were evenly divided into different sets.
The program was implemented with PyTorch.
{All model training and inferencing was performed on a server running Ubuntu 22.04 with two Intel Xeon Platinum 8375C CPUs and four NVIDIA RTX 4090 GPUs installed.}
Inferencing only involves one GPU.
All models were trained with the same training strategy (350 epochs, batch size 32, AdamW optimizer), and they used the same input resolution of $384\times384$ for search images and $192\times192$ for template images.
The learning rate was set to 3e-4 with a backbone learning rate of 3e-5, where both of them were dropped by a factor of 10 after 200 epochs.
{The backbone is trained together with the network but adopts a smaller learning rate.}
Scaling, blur, and position shifting were adopted as image augmentation.
Gaussian noise was added to the historical motion sequence.
{The augmentations were performed dynamically during the training process.}

\subsection{Results}
The evaluation was conducted based on three metrics that are commonly adopted in tracking evaluation, namely area under curve (AUC) \cite{wu2013online}, precision ($\textit{P}$) \cite{muller2018trackingnet}, and normalized precision ($\textit{P}_{\textit{norm}}$) \cite{muller2018trackingnet}.
AUC reports the needle tracking success rate by evaluating the area under the Success Plot curve as defined in \cite{wu2013online}, which offers a comprehensive evaluation of tracking robustness.
Higher AUC suggests that the tracker keeps the needle tip tracked for a larger portion of the frames.
Beyond success rate by AUC, tracking error is evaluated by $\textit{P}$ and $\textit{P}_{\textit{norm}}$ \cite{muller2018trackingnet}. 
$\textit{P}$ evaluates the accuracy of predicted position measured by Center Location Error (CLE), which is the Euclidean distance between the predicted and ground truth positions.
$\textit{P}$ is generated by calculating the percentage of frames where the CLE is less than a specific threshold, which is set to a conventional value of 20 pixels.
{$\textit{P}_{\textit{norm}}$ adjusts $\textit{P}$ to account for the scale of the object by normalizing CLE with the size of the ground-truth bounding box, providing a more consistent measure across varying object sizes.
The ground-truth bounding box is defined based on the needle tip region that is visible in the ultrasound imaging.}
Higher $\textit{P}$ and $\textit{P}_{\textit{norm}}$ can be interpreted as needle tip tracking with higher accuracy.
{In addition to these three metrics}, the average tracking errors and standard deviations in millimeters were also reported to provide a better comparison in terms of physical metrics.
Several state-of-the-art natural object trackers were {retrained on the proposed dataset and} used for comparison, including state-of-the-art transformer-based tracker MixFormerV2 \cite{cui_mixformerv2_2023}, tracker with motion prompts (SwinTrack \cite{lin2022swintrack}), tracker with aggregated historical templates (STMTrack \cite{fu2021stmtrack}), classical Siamese tracker SiamRPN++ \cite{li2019siamrpn++} and its improved variants \cite{chen2022siamban,guo2020siamcar,yu2020deformable}. 
{A US-based needle tracker by Yan et al. \cite{yan2023learning} was also used as a comparison.}

As the results show in Tab.~\ref{tab_exp} and Tab.~\ref{tab_mm}, the proposed MambaXCTrack achieves SOTA performance in almost all metrics of all evaluations {while maintaining satisfactory inference speed}.
In the phantom environment with less background noise and more stable needle appearance, while most methods offer satisfactory tracking performance in the out-of-plane cases, a few trackers, including MambaXCTrack, distinguish themselves with superior performance in the in-plane cases. SwinTrack and MixFormerV2 were ranked second in the in-plane (static) and in-plane (moving) cases, respectively, while MambaXCTrack outperforms both to rank first in both cases. This achievement can be attributed to the higher modeling efficiency of the Mamba-based structure compared with the transformer architecture adopted in both SwinTrack and MixFormerV2.
MambaXCTrack also achieves a 0.22$\pm$0.14 mm tracking error and standard deviation (SD). 
This means that the estimated tip position can be maintained within the 18-gauge needle tip (with a cross-sectional radius of 0.625 mm) in almost all frames.


\begin{table*}[t]
\begin{center}
\caption{
Ablation studies of the baseline model {(MambaXCTrack)} against seven variations.
{The results of the phantom and tissue experiments are reported as mean values of the three insertion techniques.}
}
\label{tab_ab}
\setlength{\tabcolsep}{1.3mm}{
\scalebox{1.0}{
\begin{tabular}{lccc|ccc|ccc}
\toprule
\multirow{2}{*}{Method} & \multicolumn{3}{c|}{Phantom} & \multicolumn{3}{c|}{Tissue} & \multicolumn{3}{c}{Mean}\\
& \cellcolor{c2!50}AUC & \cellcolor{c2!50}$\textit{P}_{\textit{norm}}$ & \cellcolor{c2!50}$\textit{P}$ & \cellcolor{c2!50}AUC & \cellcolor{c2!50}$\textit{P}_{\textit{norm}}$ & \cellcolor{c2!50}$\textit{P}$ & \cellcolor{c2!50}AUC & \cellcolor{c2!50}$\textit{P}_{\textit{norm}}$ & \cellcolor{c2!50}$\textit{P}$ \\
\midrule
Baseline{: MambaXCTrack} & 77.3 & 96.9 & 99.9 & 63.3 & 77.7 & 91.2 & 70.8 & 87.9 & 95.9 \\
\midrule
$v_1$: w/ CIS $[z, m, x_i]$ & 77.5{\scriptsize~(+0.2)} & 96.6{\scriptsize~(-0.3)} & 99.1{\scriptsize~(-0.8)} & 62.8{\scriptsize~(-0.5)} & 78.7{\scriptsize~(+1.0)} & 89.1{\scriptsize~(-2.1)} & 70.7{\scriptsize~(-0.1)} & 88.2{\scriptsize~(+0.3)} & 94.4{\scriptsize~(-1.5)} \\
$v_2$: w/o CIS, simply cat $[m, z, x]$ & 76.7{\scriptsize~(-0.6)} & 97.1{\scriptsize~(+0.2)} & 99.1{\scriptsize~(-0.8)} & 61.6{\scriptsize~(-1.7)} & 75.9{\scriptsize~(-1.8)} & 87.9{\scriptsize~(-3.3)} & 69.7{\scriptsize~(-1.1)} & 86.8{\scriptsize~(-1.1)} & 93.9{\scriptsize~(-2.0)} \\
$v_3$: w/o SSMX-Corr, w/ ConvX-Corr  & 73.4{\scriptsize~(-3.9)} & 94.4{\scriptsize~(-2.5)} & 97.2{\scriptsize~(-2.7)} & 57.6{\scriptsize~(-5.7)} & 73.8{\scriptsize~(-3.9)} & 85.4{\scriptsize~(-5.8)} & 66.1{\scriptsize~(-4.7)} & 84.8{\scriptsize~(-3.1)} & 91.7{\scriptsize~(-4.2)} \\
\midrule
$v_4$: w/o $m$ & 71.9{\scriptsize~(-5.4)} & 92.7{\scriptsize~(-4.2)} & 95.0{\scriptsize~(-4.9)} & 58.6{\scriptsize~(-4.7)} & 76.5{\scriptsize~(-1.2)} & 87.4{\scriptsize~(-3.8)} & 65.7{\scriptsize~(-5.1)} & 85.1{\scriptsize~(-2.8)} & 91.4{\scriptsize~(-4.5)} \\
$v_5$: w/ $m$, but raw motion & 75.5{\scriptsize~(-1.8)} & 94.8{\scriptsize~(-2.1)} & 97.2{\scriptsize~(-2.7)} & 61.7{\scriptsize~(-1.6)} & 78.1{\scriptsize~(+0.4)} & 87.9{\scriptsize~(-3.3)} & 69.0{\scriptsize~(-1.8)} & 87.0{\scriptsize~(-0.9)} & 92.9{\scriptsize~(-3.0)} \\
$v_6$: w/ $m$, $T=120$ & 77.2{\scriptsize~(-0.1)} & 96.9{\scriptsize~(-0.0)} & 99.9{\scriptsize~(-0.0)} & 61.9{\scriptsize~(-1.4)} & 77.1{\scriptsize~(-0.6)} & 89.7{\scriptsize~(-1.5)} & 70.0{\scriptsize~(-0.8)} & 87.7{\scriptsize~(-0.2)} & 95.1{\scriptsize~(-0.8)} \\
$v_7$: w/ $m$, $T=30$ & 74.5{\scriptsize~(-2.8)} & 94.6{\scriptsize~(-2.3)} & 97.4{\scriptsize~(-2.5)} & 62.7{\scriptsize~(-0.6)} & 79.8{\scriptsize~(+2.1)} & 91.6{\scriptsize~(+0.4)} & 69.0{\scriptsize~(-1.8)} & 87.7{\scriptsize~(-0.2)} & 94.7{\scriptsize~(-1.2)} \\
\bottomrule
\end{tabular}
}
}
\end{center}
\vspace{-4mm}
\end{table*}

In the tissue environment that is much more challenging than the phantom, MambaXCTrack consistently achieves the best and second-best performance in nearly all evaluations. 
MambaXCTrack significantly outperforms Yan et.al. \cite{yan2023learning} by 125.5\%, 104.8\%, and 68.1\% in AUC for all three cases.
While both MambaXCTrack and the work in \cite{yan2023learning} integrate motion information to address {the} intermittent invisibility {issues}, the superior performance by MambaXCTrack is mainly attributed to longer modeling distance of SSMX-Corr than {the} transformer in \cite{yan2023learning} and better generalizability of {an} implicit low-level motion descriptor than {an} explicit motion predictor in \cite{yan2023learning}.
As shown in Tab. II, MambaXCTrack achieves a tracking error of 0.49$\pm$1.01 mm in tissue experiments, outperforming the second-best MixFormerV2 by 64.2\% and 33.1\%.
The offset of the predicted position relative to the tip ground truth is kept within a distance of the needle cross-sectional radius, better than MixFormerV2, which has an average error of 2.2 times the needle radius.

In the average evaluation across both phantom and tissue, our tracker demonstrates superior performance across all metrics. 
Specifically, it achieves an average error of 0.34$\pm$0.55 mm, indicating that, in most frames, the tracker maintains accuracy within approximately one radius (0.635 mm), with error variations staying below 0.55 mm. This represents a substantial improvement over the second-best method, MixFormerV2, with reductions of 55.3\% in average error and 34.5\% in standard deviation.

{}


\begin{figure}
\centering
\includegraphics[width=\linewidth]{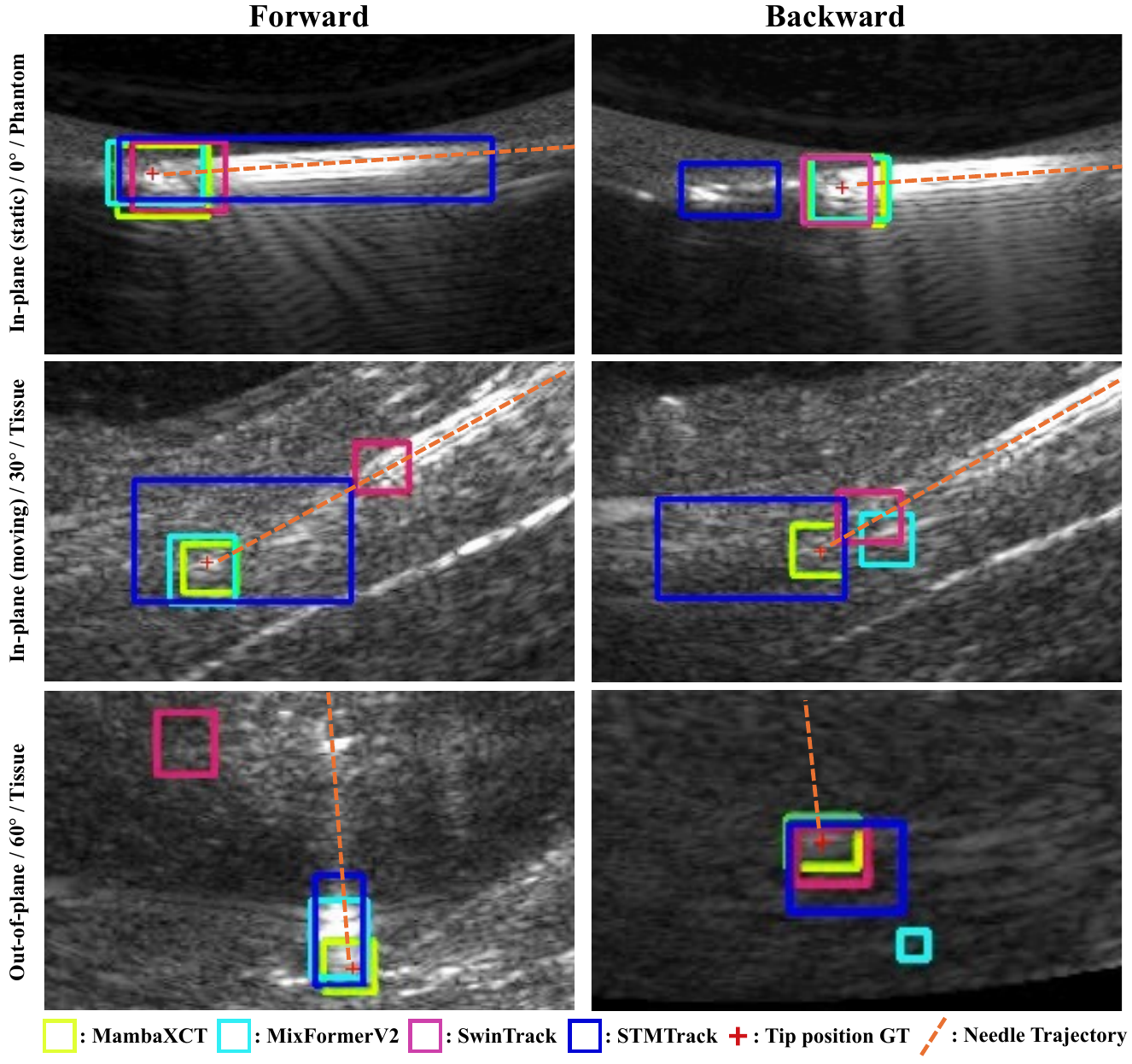}
\caption{Tracking demonstration of MambaXCTrack against three SOTA trackers in three scenarios. The view is zoomed in. The {left column} is the forward procedure, while the {right} is backward.
The {ground-truth tip position and needle trajectory} are marked. 
See more examples in the supplementary video.
}
\label{fig_demo}
\vspace{-3mm}
\end{figure}

A tracking demonstration is shown in Fig.~\ref{fig_demo} including three examples\footnote{{Additional tracking demonstrations are} provided in the supplementary video.} that demonstrate different challenges in US needle tracking.
The first example of In-plane (static) / 0$^{\circ}$ / Phantom in the first {row} shows a scenario with interference from needle traces.
MambaXCTrack overcomes the interference from needle traces, which arise from the hollow regions created by needle retraction and have a very similar appearance to the needle tip. 
This interference has led to STMTrack's failures, whereas our tracker maintains successful tracking.
For procedures done within the tissue, the distraction from noise and artifacts becomes more severe, and the needle visibility is degraded.
The second example of In-plane (moving) / 30$^{\circ}$ / Tissue demonstrates this distraction and degradation, where the background is filled with noise from high-intensity speckles caused by {variations in acoustic impedance}.
The backward procedure of this example also shows the intermittent invisible needle tip, since the imaging plane and needle tend to be misaligned in this dynamic case, where both probe and needle are moving. 
MambaXCTrack is the only one among the four that maintains successful tracking even when the tip is intermittently invisible.
A similar case of tracking {the} intermittently disappearing needle is also demonstrated in the backward of out-of-plane / 60° / Tissue, where MixFormerV2 lost tracking but MambaXCTrack was able to keep the target tracked.

{
Our experiments show that MambaXCTrack runs at 34.9 FPS, which meets the real-time requirement for ultrasound needle tracking despite being slower than the MixFormerV2 and SwinTrack. 
However, our method consistently outperforms these methods in terms of needle tracking accuracy by leveraging pixel-level global modeling with SSM, which efficiently extracts rich semantic cues from neighboring regions. 
The linear complexity of SSM enables this pixel-level global operation, which can hardly be achieved by a transformer-based method without a significant increase in computational complexity. 
Therefore, MambaXCTrack achieves an appropriate balance between inference speed and accuracy, as clinical applications demand reliability more over speed gains when real-time thresholds can be met.
}

\subsection{Ablation Study}
Extensive ablation studies were conducted regarding the baseline model {MambaXCTrack} against seven variations $v_1$ to $v_7$, with their configurations and results shown in Tab.~\ref{tab_ab}.
The first three were adopted to test the proposed SSMX-Corr and CIS.
CIS is kept in $v_1$ but $m$ is concatenated between $z$ and $x_i$, different from the baseline ($[m, z, x_i]$), to evaluate different orders of information fusion.
CIS is removed in $v_2$ and three feature maps are simply concatenated together into a sequence $[m, z, x]$.
CIS is also removed in $v_3$ and SSMX-Corr is replaced by a convolutional X-Corr \cite{li2019siamrpn++}.
The results of these variations show the effectiveness of CIS and SSMX-Corr.
Performance degradation in $v_1$ demonstrates that the tracking benefits more from motion information by concatenating $m$ at the beginning of $z$ and $x_i$.
With CIS removed ($v_2$), performance degradation is observed in almost all evaluations, showing that tracking can be improved by building pixel-wise interaction between $z$ and $x$.
Performance drop is also observed in $v_3$, especially for the tissue experiments, where noise distraction is much more severe. 
It demonstrates that tracking under challenging environments is greatly improved by learning potential distant visual cues utilizing SSMX-Corr's long-range modeling capability.

Ablation regarding the proposed low-level motion descriptor was also carried out, namely $v_4$ to $v_7$.
The motion descriptor is removed in $v_4$. $v_5$ keeps the motion descriptor but uses raw motion without preprocessing it like the baseline.
Motion descriptors that cover different time durations are tested in $v_6$ (longer time, $T=120$) and $v_7$ (shorter time, $T=30$). 
A considerable degradation is reported in $v_4$, demonstrating that tracking is greatly enhanced by learning from historical low-level motion descriptors.
Accuracy drops are also reported in almost all metrics of $v_5$, showing the improvement brought by extracting low-level descriptors from raw motion. 
The results of $v_6$ and $v_7$ demonstrate the length of baseline ($T=60$), {which covers 2 seconds}, is a more suitable time span for motion descriptors in US needle tracking. 
This is likely because moderate-duration motion provides sufficient cues while avoiding interference from long-term noise.

\subsection{Discussion}
MambaXCTrack outperforms a series of SOTA trackers including CNN-based trackers and transformer-based trackers.
Its superior performance can be attributed to two main factors.
Firstly, the SSMX-Corr with CIS effectively addresses the unique domain characteristics of US images, i.e., \textit{the local {features} from the target can become unpredictably unreliable, {but} searching for semantic cues over long distances significantly enhances tracking accuracy.}
Studies using needle shaft segmentation for guidance \cite{wijata2024needle,hui2023ultrasound} suggest that while the needle tip may occasionally become obscured, the larger needle shaft remains partially visible, providing directional information. 
Instead of direct segmentation,
SSM is employed to endow the model with long-range modeling capabilities, allowing it to learn from distant semantic features including the needle shaft.
The existing ConvX-Corr adopted in SiamRPN++ \cite{li2019siamrpn++} is a local pixel-wise operation that convolves $x$ with kernel $z$.
While it is considered effective for open-world tasks, its limited search range proves inadequate for US needle tracking, where images are often degraded by noise and artifacts, rendering local fine features unreliable. 
With SSMX-Corr, MambaXCTrack overcomes these challenges by identifying long-range semantic cues.
By further integrating CIS, SSMX-Corr retains the advantages of ConvX-Corr, gaining positional inductive bias from this convolution-like operation.

The second aspect is the implicit low-level motion descriptor. \textit{When the needle temporarily disappears, historical motion becomes more critical than visual information.}
In such cases, relying on motion information rather than vision is a more effective strategy. 
Ablation studies demonstrate that integrating historical low-level motion descriptors with an appropriate time duration enhances tracking.
SwinTrack \cite{lin2022swintrack} also incorporates motion information yet leads to worse performance than ours.
{This is partly because SwinTrack directly integrates raw motion sequences without calculating relative displacement like MambaXCTrack, which hinders the generalization.}
Additionally, its feature fusion is less efficient than our SSM, which is inherently suited for long-sequence modeling.
STMTrack \cite{fu2021stmtrack} adopts a template bank to store the historical frames, focusing on historical visual {information} rather than motion information.
However, it performs even worse than SwinTrack \cite{lin2022swintrack}.
This might be because of error accumulation due to aggregating historical US images, which contain severe noise and artifacts.
Yet historical motion suffers less from error accumulation thanks to its simpler distribution and characteristic that does not contain visual features.
In challenging environments, leveraging simpler motion information is more effective than integrating visual features.
However, while the tracker proposed by Yan et al. \cite{yan2023learning} also utilizes motion for ultrasound needle tracking, its performance is unsatisfactory.
This suboptimal performance is likely attributed to its explicit motion prediction structure, which prevents the model from being fully end-to-end, resulting in error propagation and system instability. 
Directly {predicting} future motion can also cause degraded generalizability when encountering unseen videos, which contain motion that {can be} hard to be adapted by a simple motion predictor.
On the contrary, MambaXCTrack leverages implicit motion integration to ensure an end-to-end structure with better generalization.

\section{CONCLUSIONS}
In this paper, a US needle tracker MambaXCTrack with SSMX-Corr, CIS, and a low-level motion descriptor has been introduced for US needle tracking under challenging US imaging environments with noise, artifacts, and intermittent tip visibility.
{To the best of our knowledge,} it is the first Mamba-based tracker for US needle tracking.
Experiments and ablation studies show its effectiveness.
{However, MambaXCTrack requires a high-end GPU to achieve real-time inference, hindering deployment on low-end hardware. In future work, improvements will be made for better efficiency, including network pruning, model distillation, parallel optimization, etc.}
More experiments on manual needle insertion will also be conducted.
{The code implementation of the proposed method will be made publicly available in future work once ongoing developments are completed.}






\bibliographystyle{ieeetr}
\bibliography{reference}

\end{document}